\documentclass[10pt,twocolumn,letterpaper]{article}

\usepackage{cvpr}              
\usepackage{times}
\usepackage{graphicx}
\usepackage{comment}
\usepackage{amsmath,amssymb,bm} 
\usepackage{color}
\usepackage{epsfig}
\usepackage{graphicx}
\usepackage{amsmath}
\usepackage{amssymb}
\usepackage[T1]{fontenc}
\usepackage{multirow}
\usepackage{caption}
\usepackage{mathrsfs}
\usepackage{url}
\usepackage{algorithm}
\usepackage{algorithmic}

\def\bs{\bm}

\DeclareMathOperator*{\argmax}{arg\,max}
\DeclareMathOperator*{\argmin}{arg\,min}

\usepackage[dvipsnames]{xcolor}

\definecolor{cvprblue}{rgb}{0.21,0.49,0.74}
\usepackage[pagebackref,breaklinks,colorlinks,citecolor=cvprblue]{hyperref}

\def\paperID{136} 
\def\confName{3DV}
\def\confYear{2024}

\title{A Hybrid Generative and Discriminative PointNet on Unordered Point Sets}

\author{Yang Ye\\
Department of Computer Science\\
Georgia State University\\
Atlanta, GA 30302, USA\\
{\tt\small yye10@student.gsu.edu}
\and
Shihao Ji\\
Department of Computer Science\\
Georgia State University\\
Atlanta, GA 30302, USA \\
{\tt\small sji@gsu.edu}
}

\begin{document}

\maketitle
\begin{abstract}

As point cloud provides a natural and flexible representation usable in myriad applications (e.g., robotics and self-driving cars), the ability to synthesize point clouds for analysis becomes crucial. Recently, Xie et al.~\cite{xie2021generative} propose a generative model for unordered point sets in the form of an energy-based model (EBM). Despite the model achieving an impressive performance for point cloud generation, one separate model needs to be trained for each category to capture the complex point set distributions. Besides, their method is unable to classify point clouds directly and requires additional fine-tuning for classification. One interesting question is: Can we train a single network for a hybrid generative and discriminative model of point clouds? A similar question has recently been answered in the affirmative for images, introducing the framework of Joint Energy-based Model (JEM)~\cite{jem,jempp}, which achieves high performance in image classification and generation simultaneously. This paper proposes GDPNet, the first hybrid Generative and Discriminative PointNet that extends JEM for point cloud classification and generation. Our GDPNet retains strong discriminative power of modern PointNet classifiers~\cite{qi2017pointnet}, while generating point cloud samples rivaling state-of-the-art generative approaches. 

\end{abstract}

\begin{figure}[t]
\centerline{
\includegraphics[width =6cm]{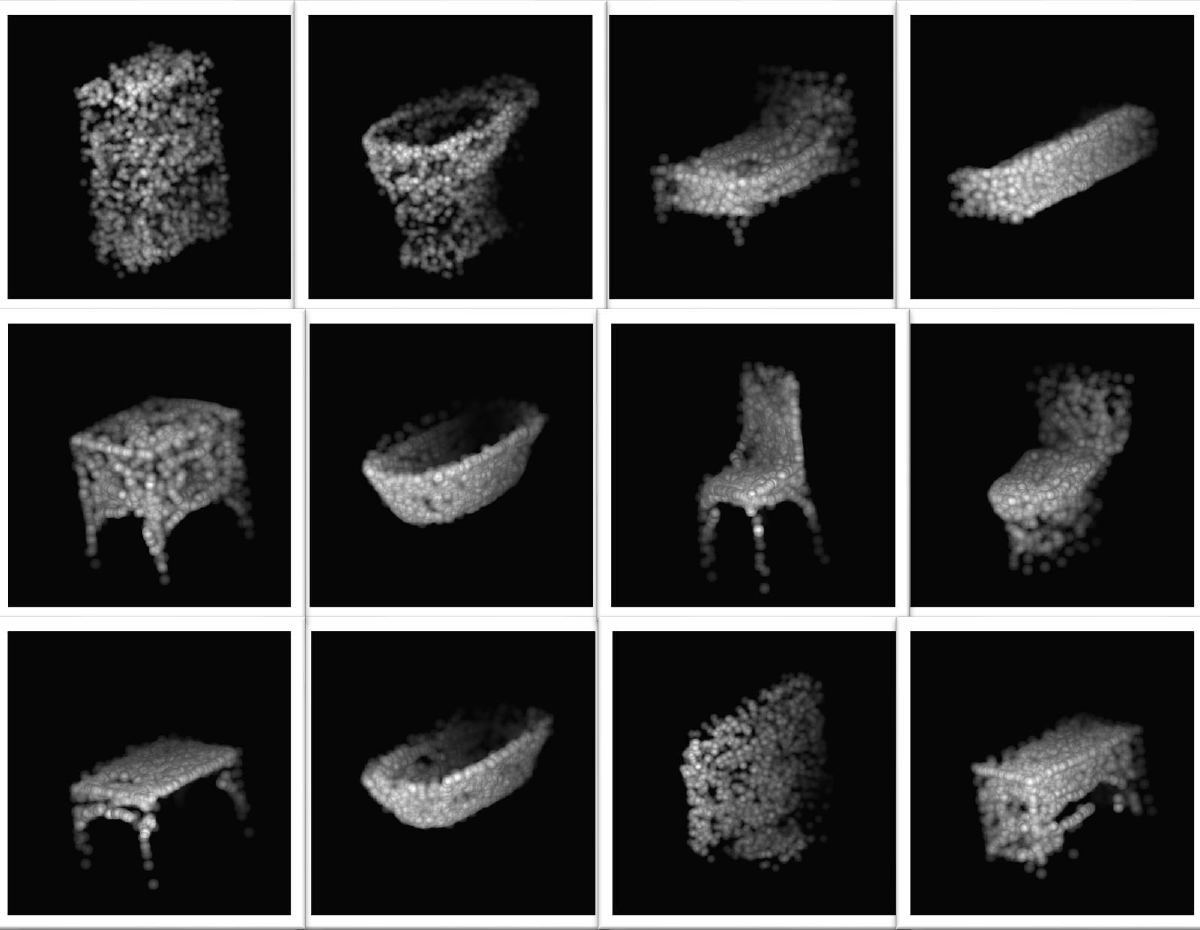}}
\caption{GDPNet performs point cloud classification and generation with a single network. It can generate 10 categories of point clouds, while achieving a 92.8\% classification accuracy on ModelNet10. Sample point clouds generated by GDPNet are provided above.} 
\label{fig:samples_all}
\end{figure}

\section{Introduction}

With the rapid development of 3D sensing devices (e.g., LiDAR and RGB-D camera), a huge amount of point cloud data are collected in the area of robotics, autonomous driving and virtual reality~\cite{Nchter2008TowardsSM,KITTI,AR}. A 3D point cloud, composed of the raw coordinates of scanned points in 3D space, is an accurate representation of an object or shape and plays a key role in the perception of the surrounding environment. In recent years, a myriad of processing methods~\cite{qi2017PointNetplusplus, yu2018pu-net, li2018pointcnn, qi2019deep} have been proposed for efficient point cloud analysis, and their performances in applications, such as 3D point cloud classification~\cite{qi2017PointNetplusplus, li2018pointcnn, thomas2019kpconv, wu2019pointconv}, semantic segmentation~\cite{li2018so-net, su2018splatnet, liu2019relation, wang2019dynamic,wang2018sgpn,landrieu2018large} and reconstruction~\cite{achlioptas2018learning, yang2018folding, han2019multi, zhao20193dpoint}, have been improved significantly. Despite the significant progress of discriminative models for point cloud classification and segmentation, the research of generative models for point clouds is still far behind the discriminative ones. Learning generative models for point clouds is crucial to characterize the data distribution and analyze point clouds, which lays the foundation for various tasks such as shape completion, upsampling, synthesis and data augmentation. Although generative models such as variational auto-encoders (VAEs) ~\cite{VAE} and generative adversarial networks (GANs) ~\cite{GAN}  have shown great success in 2D image generation, it is challenging to extend these well-established methods to unordered 3D point sets. Images are structured data on 2D grids, while point clouds lie in irregular 3D space with variable densities. Existing methods for point could generation are mainly based on volumetric data, e.g., 3D GAN~\cite{wu2016learning}, Generative VoxelNet~\cite{xie2018learning,Xie2020GenerativeVL}, 3D-INN~\cite{huang20193d}, PointGrow~\cite{sun2020pointgrow}, etc. While remarkable progress has been made, these methods have a few inherent limitations for point cloud generation. For instance, the training procedure of GAN-based approaches~\cite{wu2016learning} is rather unstable due to the adversarial losses, and the auto-regressive models~\cite{sun2020pointgrow} assume an order of point generation, which is unnatural for orderless point cloud generation and restricts the modeling flexibility.

Energy-based models (EBMs)~\cite{zhu1998filters,lecun2006tutorial} is a family of probabilistic generative models that can explicitly characterize the data distribution by learning an energy function that assigns lower values to observed data and higher values to unobserved ones. Besides, the training of EBMs can be much more stable in contrast to GANs by unifying representation and generation in one single model optimized by the maximum likelihood principle. Successful applications of EBMs include generations of images~\cite{xie2016theory,XieLuGao20}, videos~\cite{xie2017synthesizing,xie2019learning}, 3D volumetric shapes~\cite{xie2018learning,Xie2020GenerativeVL}, texts~\cite{deng2020residual}, molecules~\cite{ingraham2018learning} as well as image-to-image translation~\cite{xie2021cooperative} and out-of-distribution detection~\cite{liu2020energy_ood}. Recently, Xie et al.~\cite{xie2021generative} propose GPointNet, an EBM for unordered point cloud generation. Unlike models that leverage an encoder-decoder architecture for generation, GPointNet~\cite{xie2021generative} does not rely on either an auxiliary network or hand-crafted distance metrics to train the model. By incorporating PointNet~\cite{qi2017pointnet}, GPointNet extracts the features for each point independently and aggregates the point features from the whole point cloud into an energy scalar. The "fake" examples are then generated by the Langevin dynamics sampling~\cite{welling2011bayesian}, and the model parameters are updated based on the energy difference between the "fake" examples and the "real" observed examples in order to match the "fake" examples to the “real” ones in terms of some permutation-invariant statistical properties enabled by the energy function. Despite the model achieving an impressive performance for point cloud generation, one model needs to be trained separately for each category without sharing statistical regularities among different point cloud categories, e.g., structural smoothness and point density transition. Besides, their method is unable to classify point clouds directly and requires additional fine-tuning with an SVM classifier for classification. 

In this paper, we propose GDPNet, the first hybrid Generative and Discriminative PointNet for point cloud classification and generation with a single network. Our design follows the framework of joint energy-based model (JEM)~\cite{jem}, which reinterprets CNN softmax classifier as an EBM for image classification and generation. Instead, we extend JEM for point cloud classification and generation based on a modern PointNet classifier~\cite{qi2017pointnet}. The direct extension of JEM to PointNet, however, does not perform well as manifested by a classification accuracy gap to the standard classifier and a generation quality gap to the state-of-the-art generative approaches. We therefore investigate training techniques to bridge both gaps of GDPNet. We leverage the Sharpness-Aware Minimization (SAM)~\cite{sam2021} to improve the generalization of GDPNet (Section~\ref{sec:sam}). We further demonstrate that the smoothness of the activation function can improve the training stability and synthesis quality of GDPNet significantly. As a result, our GDPNet retains strong discriminative power of modern PointNet classifier, while generating point cloud samples rivaling state-of-the-art generative approaches. More importantly, GDPNet yields one single model for classification and generation for all point cloud categories without resorting to a dedicated model for each category or additional fine-tuning step for classification. Example point clouds generated by GDPNet are provided in Figure~\ref{fig:samples_all}.



\section{Related Work}

\subsection{Deep Learning on Point Clouds}

Following the breakthrough results of CNNs in 2D image processing tasks~\cite{AlexNet,he2016deep}, there has been a strong interest in adapting such methods to 3D geometric data. Compared to 2D images, point cloud data are sparse, unordered and locality-sensitive, making it non-trivial to adapt CNNs to point cloud processing. Early attempts focus on regular representations of the data in the form of 3D voxels~\cite{wu2015modelnet, qi2016volumetric}. These methods quantize point clouds into regular voxels in 3D space with a predefined resolution and then apply volumetric convolution. Recently, new designs of local aggregators over point clouds are proposed to improve the efficiency of point cloud processing and reduce the loss of details~\cite{qi2017pointnet,qi2017PointNetplusplus,wang2019dynamic}. PointNet~\cite{qi2017pointnet} is a pioneer in deep architecture design that directly processes point clouds for classification and semantic segmentation by a shared multi-layer perception (MLP) and a max-pooling layer. However, it treats each point independently and ignores the geometric relationships among them, and thus only local features are extracted. PointNet++~\cite{qi2017PointNetplusplus} further introduces a hierarchical aggregation of point features to extract global features. In later works, DGCNN~\cite{wang2019dynamic} proposes an effective EdgeConv that encodes the point relationships as edge features to better capture local geometric features, while still maintaining permutation-invariance. Our GDPNet reinterprets PointNet classifier as an EBM and empowers it for point cloud generation while retaining its strong discriminative power.


\subsection{Point Cloud Generation}
Since point clouds lie in irregular 3D space with variable densities, early point cloud generation methods~\cite{achlioptas2018learning, gadelha2018multiresolution} convert the point cloud generation into a matrix generation problem. They take advantage of the power of well-established frameworks of variational auto-encoders (VAEs)~\cite{VAE} and generative adversarial networks (GANs)~\cite{GAN} to train generative models with hand-crafted distance metrics, such as Chamfer distance or earth mover’s distance, to measure the dissimilarity of two point clouds. The main defect of these methods is that they are restricted to generating point clouds with a fixed number of points and lack the property of permutation-invariance. FoldingNet~\cite{yang2018folding}  and AtlasNet~\cite{groueix2018papier} learn a mapping that deforms the 2D patches into 3D shapes of point clouds to generate an arbitrary number of points, while being permutation-invariant. On the other hand, point clouds can also be regarded as samples from a point distribution and the maximum likelihood principle can be utilized for point cloud generation. PointFlow~\cite{yang2019pointflow} employs continuous normalizing flows~\cite{grathwohl2018ffjord} to model the distribution of points. The invertibility of normalizing flows enables the computation of the likelihood during training and the variational inference is adopted for model training. PointGrow~\cite{sun2020pointgrow} is an auto-regressive model that dynamically aggregates long-range dependencies among points for point cloud generation. ShapeGF~\cite{cai2020learning} proposes a score-matching energy-based model to represent the distribution of points. Luo and Hu~\cite{luo2021diffusion} view points in a point cloud as particles in a thermodynamic system that diffuse from the original distribution to a noise distribution and leverage the reverse diffusion Markov chain to model the distribution of points. GPointNet~\cite{xie2021generative} explicitly models this distribution as an EBM and learns the model by the maximum likelihood estimation. However, all these methods focus on point cloud generation and cannot perform classification at the same time. To the best of our knowledge, our GDPNet is the first hybrid generative and discriminative model for point cloud classification and generation with a single network.

\subsection{Flat Minima and Generalization}

A great number of prior works have investigated the relationship between the flatness of local minima and the generalization of learned models~\cite{losslandscape,keskar2016large,wei2020implicit,vitsam,sam2021,asam2021}. Now it is widely accepted and empirically verified that flat minima tend to give better generalization performance. Based on these observations, several recent regularization techniques are proposed to search for the flat minima of loss landscapes~\cite{wei2020implicit,vitsam,sam2021,asam2021}. Among them, the Sharpness-Aware Minimization (SAM)~\cite{sam2021} is a recently introduced optimizer that demonstrates promising performance across all kinds of models and tasks, such as ResNet~\cite{resnet16}, Vision Transformer ~\cite{vitsam} and Language Models~\cite{samlm}. Furthermore, score matching-based methods~\cite{hyvarinen2005estimation,swersky2011autoencoders,song2019generative,song2020score} also explore the behavior of flat minima in generative models and learn unnormalized statistical models by matching the gradient of the log probability density of model distribution to that of data distribution. Our GDPNet incorporates SAM to promote the energy landscape smoothness and thus improves the generalization of trained EBMs.

\section{The Proposed Method}

We first introduce the Joint Energy-based Models (JEM)~\cite{jem} and discuss its extension to GDPNet, the first hybrid generative and discriminative model for point clouds. We then present the Sharpness-Aware-Minimization (SAM)~\cite{sam2021} and its integration to GDPNet to improve the generalization of trained EBMs.

\subsection{Joint Energy-based Models for Point Clouds}

Let $\bs{X} = \{\bs{x}_i\in\mathcal{R}^3\}_{i = 1}^n$ denote a point cloud that contains $n$ points, with $\bs{x}_i$ representing the 3D coordinates of point $i$. Following the framework of Energy-Based Models (EBMs)~\cite{zhu1998filters,lecun2006tutorial}, we define the probability density function of a point cloud $\bs{X}$ explicitly as 
\begin{equation}\label{eq:ebm_define}
  p_{\bs{\theta}}(\bs{X})=\frac{\exp \left(-E_{\bs{\theta}}(\bs{X})\right)}{Z(\bs{\theta})},
\end{equation}
where $E_{\bs{\theta}}(\bs{X})$ is an energy function, parameterized by $\bs{\theta}$, that maps input point cloud $\bs{X}$ to a scalar, and $Z(\bs{\theta}) = \int_{\bs{X}} \exp \left(-E_{\bs{\theta}}(\bs{X})\right)$ is the normalizing constant w.r.t. $\bs{X}$ (also known as the partition function). Ideally, the energy function should assign low energy values to the samples drawn from data distribution, and high values otherwise. Since point cloud $\bs{X}$ is a set of unordered points, the energy function, $E_{\bs{\theta}}(\bs{X})$, defined on a point set needs to be invariant to the permutation of points in the set $\bs{X}$. We follow the design of PointNet~\cite{qi2017pointnet} to employ a shared multi-layer perception (MLP) for each point in the set $\bs{X}$, followed by an average-pooling layer, to approximate a continuous set function to process the unordered point sets.

The maximum likelihood estimate (MLE) can be used to estimate parameters $\bs{\theta}$ of $E_{\bs{\theta}}(\bs{X})$. However, since the partition function $Z(\bs{\theta})$ is intractable, the MLE of $\bs{\theta}$ is not straightforward. Specifically, the derivative of the log-likelihood of $\bs{X}$ w.r.t. $\bs{\theta}$ can be expressed as
\begin{align}\label{eq:ml}
  \!\!\!\!\frac{\partial\!\log p_{\bs{\theta}}(\bs{X})}{\partial\bs{\theta}}\!\!=\!   \mathbb{E}_{p_{\bs{\theta}}(\bs{X})} \!\! \left[ \frac{\partial\!E_{\bs{\theta}}(\bs{X})}{\partial\bs{\theta}} \!\right]
  \!\!-\!  \mathbb{E}_{p_d(\bs{X})} \!\! \left[ \frac{\partial\!E_{\bs{\theta}}(\bs{X})}{\partial\bs{\theta}} \!\right],
\end{align}
where $p_d(\bs{X})$ is the real data distribution (i.e., training dataset), and $p_{\bs{\theta}}(\bs{X})$ is the estimated probability density function (\ref{eq:ebm_define}), sampling from which is challenging due to the intractable $Z(\bs{\theta})$. 

Prior works have developed a number of methods to sample from $p_{\bs{\theta}}(\bs{X})$ efficiently, such as MCMC and Gibbs sampling~\cite{hinton2002cd}. To speed up the sampling process further, recently Stochastic Gradient Langevin Dynamics (SGLD)~\cite{welling2011bayesian} has been employed to sample from $p_{\bs{\theta}}(\bs{X})$ by utilizing the gradient information~\cite{nijkamp2019learning,du2019implicit,jem}. Specifically, to sample from $p_{\bs{\theta}}(\bs{X})$, SGLD follows
\begin{align}\label{eq:sgld}
    &\bs{X}^0\sim p_0(\bs{X}),
    &\bs{X}^{t+1} = \bs{X}^t-\frac{\alpha}{2} \frac{\partial
    E_{\bs{\theta}}(\bs{X}^t)}{\partial \bs{X}^t} + \alpha\bs{\epsilon}^t,
\end{align}
where $\bs{\epsilon}^t$ is random noise that is sampled from a unit Gaussian distribution $\mathcal{N} (\bs{0},\bs{1})$, and $p_0(\bs{X})$ is typically a uniform distribution over $[-1,1]$, whose samples are refined via a noisy gradient decent with step-size $\alpha$ over a sampling chain. 

In order to train a hybrid generative and discriminative model, Joint Energy-based Model (JEM)~\cite{jem} reinterprets the standard softmax classifier as an EBM. In particular, the logits $f_{\bs{\theta}}(\bs{X})[y]$ from a standard softmax classifier can be considered as an energy function over $(\bs{X}, y)$, where $y$ is class label, and thus the joint density function of $(\bs{X}, y)$ can be expressed as $p_{\bs{\theta}}(\bs{X}, y)=e^{f_{\bs{\theta}}(\bs{X})[y]} / Z(\bs{\theta})$, where $Z(\bs{\theta})$ is an unknown normalizing constant (regardless of $\bs{X}$ or $y$). Then the density of $\bs{X}$ can be derived by marginalizing over $y$: $p_{\bs{\theta}}(\bs{X})=\sum_{y} p_{\bs{\theta}}(\bs{X}, y) = \sum_{y} e^{f_{\bs{\theta}}(\bs{X})\left[y\right]} / Z(\bs{\theta})$. Subsequently, the corresponding energy function of $\bs{X}$ can be identified as 
\begin{equation}\label{eq:jem_ex}
    E_{\bs{\theta}} (\bs{X})\!=\!-\log\! \sum_{y}\!\exp({f_{\bs{\theta}}(\bs{X})\left[y\right]})\!=\!-\text{LSE}( f_{\bs{\theta}}(\bs{X})),
\end{equation}
where $\text{LSE}(\cdot)$ denotes the Log-Sum-Exp function.

\begin{figure}[t]
\centerline{
\includegraphics[width = 7.8cm]{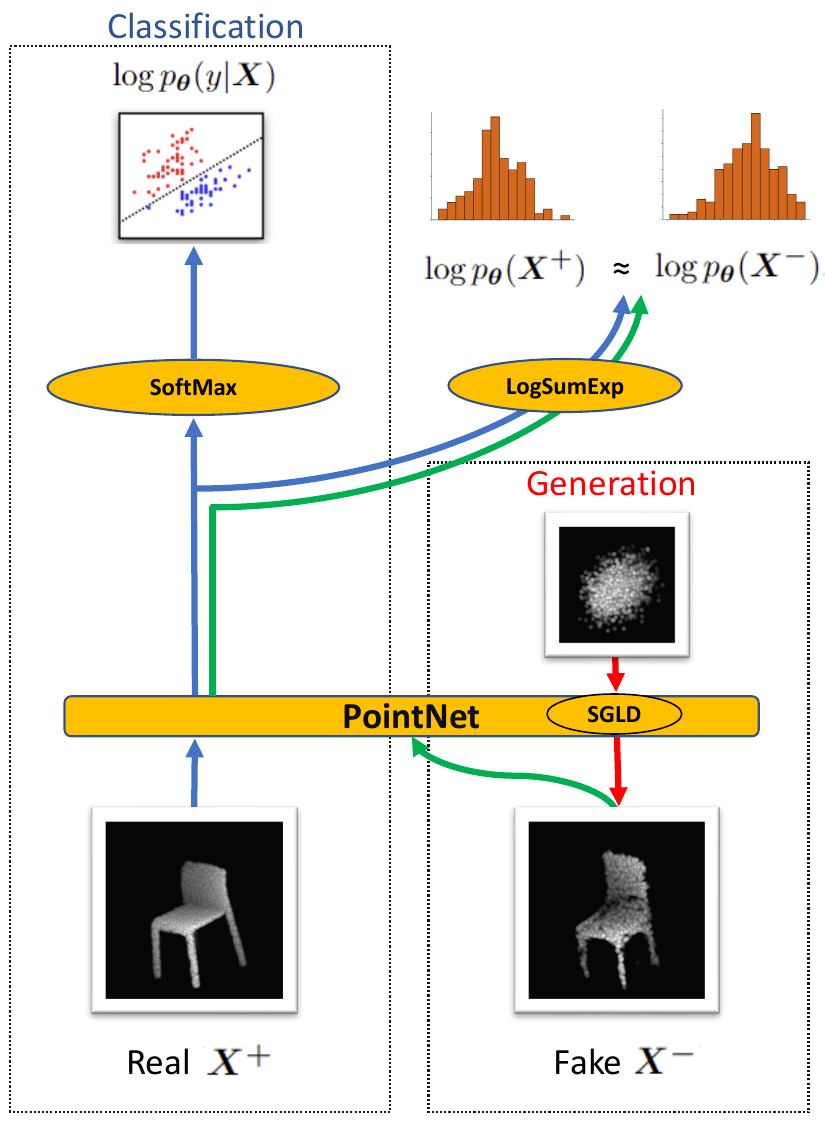}}
\caption{Overview of GDPNet architecture. Following the design of JEM~\cite{jem}, PointNet is leveraged for unordered point set feature extraction, and the LogSumExp(·) of the logits from the softmax classifier can be re-used to define an energy function of point cloud $\bs{X}$, which leads to a hybrid generative and discriminative model with the fake samples generated from the SGLD sampling. The model is optimized to perform the classification and maximize the energy difference between fake and real samples.}
\label{fig:overview}
\end{figure}

To optimize the model parameter $\bs{\theta}$, JEM maximizes the logarithm of joint density function $p_{\bs{\theta}}(\bs{X},y)$:
\begin{equation}\label{eq:jem_loss}
  \log p_{\bs{\theta}}(\bs{X}, y) = \log p_{\bs{\theta}}(y|\bs{X}) + \log p_{\bs{\theta}}(\bs{X}),
\end{equation}
where the first term denotes the cross-entropy objective for classification, and the second term can be optimized by the maximum likelihood learning of EBM as shown in Eq.~(\ref{eq:ml}). Sharing the same objective function~(\ref{eq:jem_loss}) with JEM, GDPNet optimizes a single PointNet backbone for point cloud classification and generation. The overview of GDPNet is depicted in Figure~\ref{fig:overview}.


\subsection{Sharpness-Aware Minimization}\label{sec:sam}

The direct extension of JEM to PointNet above does not perform very well as manifested in our empirical studies (Tables~\ref{tab:comparison},~\ref{tab:classification}). In general, we notice two performance gaps of GDPNet as compared to the standard PointNet classifier and state-of-the-art generative approaches, i.e., a classification accuracy gap and a generation quality gap. We therefore investigate training techniques to bridge both gaps of GDPNet. In particular, we leverage the Sharpness-Aware Minimization (SAM)~\cite{sam2021} to improve the generalization of GDPNet.

SAM~\cite{sam2021} is a recently proposed optimization method that searches for model parameters $\bs{\theta}$ whose entire neighborhoods have uniformly low loss values by optimizing a minimax objective:
\begin{align}\label{eq:sam_obj}
    \min_{\bs{\theta}}\ \max_{\| \bs{\epsilon} \|_2\leq \rho} L_{train}(\bs{\theta}+ \bs{\epsilon}) + \lambda \| \bs{\theta} \|^2_2,
\end{align}
where $\rho$ is the radius of an $L_2$-ball centered at model parameter $\bs{\theta}$, and $\lambda$ is a hyperparameter for $L_2$ regularization on $\bs{\theta}$. To solve the inner maximization problem, SAM employs the Taylor expansion to develop an efficient first-order approximation to the optimal $\bs{\epsilon}^*$ as:
\begin{align}\label{eq:e_theta}
    \hat{\bs{\epsilon}}(\bs{\theta}) &= \argmax_{\|\epsilon\|_2\leq\rho}L_{train}(\bs{\theta}) + \epsilon^T\nabla_{\bs{\theta}} L_{train}(\bs{\theta})    \nonumber \\
    &= \rho\nabla_{\bs{\theta}} L_{train}(\bs{\theta})/\|\nabla_{\bs{\theta}} L_{train}(\bs{\theta})\|_2,
\end{align}
which is a scaled $L_2$ normalized gradient at the current model parameters $\bs{\theta}$. After $\hat{\bs{\epsilon}}$ is determined, SAM updates $\bs{\theta}$ based on the gradient $\nabla_{\bs{\theta}} L_{train}(\bs{\theta})|_{\bs{\theta}+\hat{\epsilon}(\bs{\theta})}+2\lambda\bs{\theta}$ at an updated parameter location $\bs{\theta}+\hat{\epsilon}$. 


We incorporate SAM into the original training pipeline of GDPNet in order to improve the generalization of trained EBMs. Specifically, instead of the traditional maximum likelihood training of objective~(\ref{eq:jem_loss}), we optimize the joint density function in a minimax objective:
\begin{align}\label{eq:sajem_obj}
    \max_{\bs{\theta}}\ \min_{\| \bs{\epsilon} \|_2\leq \rho} \log p_{(\bs{\theta}+ \bs{\epsilon})}(\bs{X},y) - \lambda \| \bs{\theta} \|^2_2.
\end{align}
For the outer maximization that involves $\log p_{\bs{\theta}}(\bs{X})$, SGLD is again used to sample from $p_{\bs{\theta}}(\bs{X})$ as in the original JEM.

\subsection{Smooth Activation Functions}\label{sec:celu}
We further study the effect of the activation function used in the energy function $E_{\bs{\theta}}(\bs{X})$. Zhao et al.~\cite{coarse2fine2021ebm} demonstrate that when data $\bs{X}$ is continuous, the smoothness of the activation function will substantially affect the Langevin sampling process (because the derivative $\frac{\partial E_{\bs{\theta}}(\bs{X})}{\partial\bs{X}}$ is inside of Eq.~\ref{eq:sgld}). Therefore, employing an activation function with continuous gradients everywhere can stabilize the sampling, while the non-smooth activation functions like ReLU~\cite{relu10} and LeakyReLU ~\cite{leakyrelu13} may cause the divergence of EBM training. From our empirical studies, we have similar observations with the details provided in the experiments.

\begin{algorithm}[t!]
\caption{GDPNet training: Given network $f_\theta$, SGLD step-size $\alpha$, SGLD noise $\sigma$, SGLD steps $K$, replay buffer $B$, reinitialization frequency $\gamma$, SAM noise bound $\rho$, and learning rate $lr$}
\label{algo:1}
\begin{algorithmic}[1]
\WHILE{not converged}
\STATE Sample $\bs{X}^+$ and $y$ from training dataset
\STATE Sample $\widehat{\bs{X}}_0 \sim B$ with probability $1-\gamma$,  else $\widehat{\bs{X}}_0 \sim p_0(\bs{X})$  
\FOR{$t = 1, 2, \cdots, K$}   
  \STATE $\widehat{\bs{X}}_t = \widehat{\bs{X}}_{t-1} - \alpha \cdot \frac{\partial E(\widehat{\bs{X}}_{t-1})}{\partial \widehat{\bs{X}}_{t-1}} + \sigma \cdot \mathcal{N}(0, I)$
\ENDFOR
\STATE $\bs{X}^- = \text{StopGrad}(\widehat{\bs{X}}_K)$
\STATE $L_{\text{gen}}(\theta) = E(\bs{X}^+) - E(\bs{X}^-)$ \label{row:cd}
\STATE $L(\bs{\theta}) = L_\text{clf}(\bs{\theta}) +  L_{\text{gen}}(\bs{\theta})$ with  $L_{\text{clf}}(\bs{\theta}) = \text{xent}(f_{\bs{\theta}}(\bs{X}^+), y)$ \label{row:full}
\STATE \# Apply SAM optimizer as follows:
\STATE Compute gradient $\nabla_{\bs{\theta}}{L(\bs{\theta})}$ of the training loss
\STATE Compute $\hat{\bs{\epsilon}}(\bs{\theta})$ with $\rho$ as in Eq.~(\ref{eq:e_theta})
\STATE Compute gradient $\bs{g} = \nabla_{\bs{\theta}}{L(\bs{\theta})}|_{\bs{\theta} + \hat{\bs{\epsilon}(\bs{\theta})}}$
\STATE Update model parameters: $\bs{\theta} = \bs{\theta} - lr \cdot  \bs{g}$
\STATE Add $\bs{X}^-$ to $B$
\ENDWHILE
\end{algorithmic}
\end{algorithm}

\subsection{Training Algorithm}
The pseudo-code of training GDPNet is provided in Algorithm~\ref{algo:1}, which follows a similar design of JEM~\cite{jem} and JEM++~\cite{jempp} with a replay buffer. For brevity, only one real sample and one generated sample are used to optimize the model parameter $\bs{\theta}$. But it is straightforward to generalize the pseudo-code to a mini-batch setting, which we use in our experiments. It is worth mentioning that we adopt the Informative Initialization in JEM++ to initialize the Markov chain from $p_0(\bs{X})$, which enables batch normalization~\cite{batchnorm15} in PointNet and plays a crucial role in the tradeoff between the number of SGLD sampling steps $K$ and overall performance, including the classification accuracy and training stability.

\section{Experiments}
We evaluate the classification and generation performance of GDPNet in this section, and compare it with standard PointNet classifier~\cite{qi2017pointnet} and state-of-the-art generative models, including PointFlow~\cite{yang2019pointflow} and GPointNet~\cite{xie2021generative}. Ablation studies are performed to illustrate the impacts of SAM and smooth activation functions on the performance of GDPNet. Our source code is provided as a part of supplementary materials.


\subsection{Experimental Setup}

Our experiments largely follow the setup of GPointNet~\cite{xie2021generative}, and evaluate GDPNet for point cloud classification and generation on ModelNet10, which is a 10-category subset of ModelNet~\cite{wu20153d}. We first create a dataset from ModelNet10 by sampling 2,048 points uniformly from the mesh surface of each object and then scale the point cloud features to the range of [-1, 1]. In contrast to GPointNet~\cite{xie2021generative}, which trains 10 models to generate point clouds for 10 different categories of ModelNet10, we train one single network to classify and generate point clouds for all 10 categories. 

For a fair comparison with GPointNet, PointNet~\cite{qi2017pointnet} is used as the backbone network of our GDPNet. As discussed earlier, PointNet is permutation-invariant and thus works well with unordered point sets. It first maps each point (i.e., 3-dimensional coordinates) of a point cloud to a 1,024-dimensional feature vector by an MLP, then leverages an average pooling layer to aggregate information from all the points to a 1,024-dimensional global feature vector to represent the point cloud. A softmax layer is then appended at the end of the network to yield the logits for the 10 categories, which are used for classification and to calculate the energy score via an $\text{LSE}(\cdot)$ operator (Eq.~\ref{eq:jem_ex}).

Furthermore, GDPNet employs SAM~\cite{sam2021} to improve the generalization of trained EBMs. In our experiments, we use Adam~\cite{adam2014} as the base optimizer for SAM with an initial learning rate of 0.01, $\beta_1 = 0.9$ and $\beta_2 = 0.999$. We set the learning rate decay multiplier to 0.2 for every 50 iterations. We adopt the informative initialization of JEM++~\cite{jempp} to initialize the Markov chain from $p_0(\bs{X})$. The reinitialization frequency $\gamma$ is set to 0.05, and the replay buffer size is set to 5000. With the informative initialization, the number of SGLD sampling steps $K$ can be reduced significantly as compared to that of GPointNet~\cite{xie2021generative}. In our experiments, we set $K=32$ with a step size $\alpha=0.05$. To mitigate the exploding gradients in the SGLD sampling, we clip the gradient values to the range of [-1, 1] at each sampling step. We run 200 epochs for training with a minibatch size of 128.

\begin{table*}[t]
	\small
	\centering
	\begin{tabular}{|l|l|c|cc|cc|}
		\hline
		\multirow{2}{*}{\rotatebox{90}{}}  & \multirow{2}{*}{Model}   &  \multirow{2}{*}{JSD ($\downarrow$)}  & \multicolumn{2}{c|}{MMD ($\downarrow$)}        & \multicolumn{2}{c|}{Coverage ($\uparrow$)}    \\ \cline{4-7} 
		&                       &             & CD            & EMD            & CD             & EMD            \\\hline \hline

        \multirow{6}{*}{\rotatebox{90}{night stand}} & r-GAN        & 2.679            & 1.163            & 2.394            & 50.00          & 38.37          \\
                                    & l-GAN        & 1.000            & 0.746            & 1.563            & 44.19          & 39.53          \\
                                    & PointFlow    & \textbf{0.240} & 0.888 & 1.451 & 55.81 & 39.53             \\
                                    & GPointNet         & 0.590            & \textbf{0.692}   & \textbf{1.148}   & \textbf{59.30} & \textbf{61.63} \\
                                    & Ours         & 0.3771           & 0.886  & 1.369   & 54.83 & 59.30\\\cline{2-7}
                                    & Training Set & 0.263          & 0.793          & 1.096          & 60.40           & 52.32          \\\hline
        \multirow{6}{*}{\rotatebox{90}{toilet}}       & r-GAN        & 3.180            & 2.995            & 2.891            & 17.00          & 16.00             \\
                                    & l-GAN        & 1.253            & 1.258            & 1.481            & 21.00          & 28.00             \\
                                    & PointFlow    & \textbf{0.362} & 0.965 & 1.513 & 39.00 & 33.00             \\
                                    & GPointNet         & 0.386          & \textbf{0.816} & \textbf{1.265} & \textbf{44.00}    & 37.00    \\
                                    & Ours         & 0.578          & 0.867 & 1.314 & 43.00    & \textbf{42.00}    \\\cline{2-7}
                                    & Training Set & 0.249          & 0.823          & 1.116           & 48.00         & 51.00          \\\hline
        \multirow{6}{*}{\rotatebox{90}{monitor}}      & r-GAN        & 2.936          & 1.524          & 2.021          & 21.00             & 24.00             \\
                                    & l-GAN        & 1.653          & 0.915          & 1.349          & 28.00             & 27.00             \\
                                    & PointFlow    & \textbf{0.326} & 0.831 & 1.288 & 37.00 & 32.00           \\
                                    & GPointNet         & 0.780            & 0.803 & 1.213 & 40.00  & 38.00    \\
                                    & Ours         & 0.434            & \textbf{0.535} & \textbf{1.029} & \textbf{52.00}             & \textbf{46.00}    \\\cline{2-7}
                                    & Training Set & 0.283          & 0.554          & 0.938           & 48.00         & 53.00          \\\hline
        \multirow{6}{*}{\rotatebox{90}{chair}}        & r-GAN        & 2.772          & 1.709          & 2.164          & 23.00             & 28.00             \\
                                    & l-GAN        & 1.358          & 1.419          & 1.480            & 23.00             & 26.00             \\
                                    & PointFlow    & \textbf{0.278} & 0.965 & 1.322 & 42.00 & 51.00          \\
                                    & GPointNet         & 0.563          & \textbf{0.889} & \textbf{1.280} & \textbf{56.00}    & \textbf{57.00}    \\
                                    & Ours         & 0.387          & 0.909 & 1.361 & 44.00    & 50.00    \\\cline{2-7}
                                    & Training Set & 0.365          & 0.858          & 1.190          & 54.00             & 59.00          \\\hline
        \multirow{6}{*}{\rotatebox{90}{bathtub}}      & r-GAN        & 3.014          & 2.478          & 2.536          & 26.00             & 30.00             \\
                                    & l-GAN        & 0.928          & 0.865          & 1.324           & 32.00             & 38.00             \\
                                    & PointFlow    & \textbf{0.350} & \textbf{0.593} & 1.320 & 50.00 & 44.00             \\
                                    & GPointNet         & 0.460          & 0.660          & \textbf{1.108} & \textbf{58.00} & \textbf{50.00}    \\
                                    & Ours         & 0.490          & 0.647          & 1.103 & 54.00 & \textbf{50.00}    \\\cline{2-7}
                                    & Training Set & 0.344          & 0.652           & 0.980          & 56.00             & 52.00          \\\hline
	\end{tabular}
	\hspace{1mm}
	\begin{tabular}{|l|l|c|cc|cc|}
		\hline
		\multirow{6}{*}{\rotatebox{90}{}}  & \multirow{2}{*}{Model}   &  \multirow{2}{*}{JSD ($\downarrow$)}  & \multicolumn{2}{c|}{MMD ($\downarrow$)}        & \multicolumn{2}{c|}{Coverage ($\uparrow$)}    \\ \cline{4-7} 
		&                       &             & CD            & EMD            & CD             & EMD            \\\hline \hline
		        \multirow{5}{*}{\rotatebox{90}{sofa}}         & r-GAN        & 1.866          & 2.037         & 2.247          & 13.00             & 23.00             \\
                                    & l-GAN        & 0.681          & 0.631          & \textbf{1.028} & 43.00    & 44.00             \\
                                    & PointFlow    & \textbf{0.244} & 0.585 & 1.313 & 34.00 & 33.00          \\
                                    & GPointNet         & 0.647          & \textbf{0.547} & 1.089          & 39.00             & 45.00    \\
                                    & Ours         & 0.275          & 0.576 & 1.104          & \textbf{45.00}            & \textbf{46.00}    \\\cline{2-7}
                                    & Training Set & 0.185          & 0.467          & 0.904          & 56.00             & 56.00             \\\hline
        \multirow{6}{*}{\rotatebox{90}{bed}}          & r-GAN        & 1.973          & 1.250          & 2.441          & 27.00             & 21.00             \\
                                    & l-GAN        & 0.646          & \textbf{0.539}          & \textbf{0.992}          & 48.00             & 44.00             \\
                                    & PointFlow    & \textbf{0.219} & 0.544 & 1.230 & \textbf{50.00} & 35.00          \\
                                    & GPointNet         & 0.461          & 0.552          & 1.004 & \textbf{50.00}    & \textbf{50.00}    \\
                                    & Ours         & 0.240          & 0.540          & 1.088 & 45.00    & 41.00    \\\cline{2-7}
                                    & Training Set & 0.169          & 0.516          & 0.927          & 57.00          & 55.00             \\\hline
        \multirow{6}{*}{\rotatebox{90}{table}}        & r-GAN        & 3.801          & 3.714          & 2.625          & 8.00              & 14.00             \\
                                    & l-GAN        & 4.254          & 1.232           & 2.166          & 14.00             & 9.00              \\
                                    & PointFlow    & 1.044 & 1.630 & 1.535 & 16.00 & 29.00             \\
                                    &   GPointNet       & 0.869 & \textbf{0.640} & \textbf{1.000} & \textbf{44.00}    & \textbf{37.00}    \\
                                    & Ours         & \textbf{0.761} & 1.085 & 1.299 & 38.00    & 33.00    \\\cline{2-7}
                                    & Training Set & 0.703          & 1.218           & 1.182          & 31.00             & 38.00          \\\hline
        \multirow{6}{*}{\rotatebox{90}{desk}}        & r-GAN        & 3.575          & 2.712          & 3.678          & 22.09          & 22.09          \\
                                    & l-GAN        & 2.233          & 1.139 & 2.345          & 38.37          & 25.58          \\
                                    & PointFlow    & 0.327 & 1.254 & 1.548 & 38.37 & 46.51             \\
                                    & GPointNet         & 0.454          & 1.223          & 1.567          & \textbf{56.98} & \textbf{52.33} \\
                                    
                                    & Ours         & 0.512          & \textbf{1.077}          & \textbf{1.486}         & 55.81 & 50.51 \\\cline{2-7}
                                    & Training Set & 0.329          & 1.055          & 1.332          & 53.48          & 50.00             \\\hline
        \multirow{6}{*}{\rotatebox{90}{dresser}}      & r-GAN        & 1.726          & 1.299          & 1.675           & 36.05          & 30.23          \\
                                    & l-GAN        & 0.648          & 0.642          & 1.010          & 45.35          & 43.02          \\
                                    & PointFlow    & \textbf{0.270} & 0.715 & 1.349 & 46.51 & 37.21          \\
                                    & GPointNet         & 0.457          & \textbf{0.485}  & \textbf{0.988} & \textbf{53.49} & \textbf{52.33} \\
                                    
                                    & Ours         & 0.440          & 0.708  & 1.125 & 48.88 & 49.53 \\\cline{2-7}
                                    & Training Set & 0.215          & 0.551          & 0.882           & 56.98           & 54.65         \\\hline
	\end{tabular}
	\caption{Qualities of point cloud synthesis on ModelNet10 from different methods. In contrast to the state-of-the-art generative approaches, GDPNet only trains a single network to generate all the 10 categories of ModelNet10. $\downarrow$: the lower the better; $\uparrow$: the higher the better. MMD-CD scores are multiplied by 100; MMD-EMD scores and JSDs are multiplied by 10.}
	\label{tab:generation}  
\end{table*}

\subsection{Evaluation Metrics}

We adopt three evaluation metrics: Jensen-Shannon Divergence (JSD), Coverage (COV) and Minimum Matching Distance (MMD) to evaluate the quality of generated point clouds. These metrics are commonly used in prior works~\cite{achlioptas2018learning,yang2019pointflow,xie2021generative} for point cloud quality evaluation. When evaluating COV and MMD, we use Chamfer Distance (CD) and Earth Mover's Distance (EMD) to measure the dissimilarity between two point clouds, which are defined formally as follows:
\begin{align*}
&\text{CD}(X,Y) = \sum_{x\in X} \min_{y\in Y} \|x-y\|_2^2 + \sum_{y\in Y}\min_{x \in X} \|x-y\|_2^2, \\
&\text{EMD}(X,Y) = \min_{\phi: X\to Y} \sum_{x\in X} \|x-\phi(x)\|_2,
\end{align*}
where $X$ and $Y$ are two point clouds with the same number of points and $\phi$ is a bijection between them.

\textbf{Jensen-Shannon Divergence (JSD)} is a symmetrized Kullback-Leibler divergence between two marginal point distributions:
	\begin{align*}
	\text{JSD}(P_g,P_r) = \frac{1}{2}D_{KL}(P_r||M) + \frac{1}{2}D_{KL}(P_g||M)\,,
	\end{align*}
where $M=\frac{1}{2}(P_r + P_g)$, $P_r$ and $P_g$ are marginal distributions of points in the reference and generated sets, approximated by discretizing the space into $28^3$ voxels and assigning each point to one of them.	

\textbf{Coverage (COV)} measures the fraction of point clouds in the reference set that are matched to at least one point cloud in the generated set. For each point cloud in the generated set, its nearest neighbor in the reference set is marked as a match:
	\begin{align*}
	\text{COV}(S_g, S_r) = \frac{|\{\argmin_{Y \in S_r} D(X,Y) | X \in S_g \}|}{|S_r|},
	\end{align*}
where $D(\cdot, \cdot)$ can be either CD or EMD.

\textbf{Minimum Matching Distance (MMD)} is proposed to complement coverage to measure the quality of generated point clouds. For each point cloud in the reference set, the distance to its nearest neighbor in the generated set is computed and averaged:
	\begin{equation}
	\text{MMD}(S_g, S_r) = \frac{1}{|S_r|}\sum_{Y\in S_r} \min_{X\in S_g} D(X,Y).\nonumber
	\end{equation}

\begin{figure}[t]
\centerline{
\includegraphics[width = 8cm]{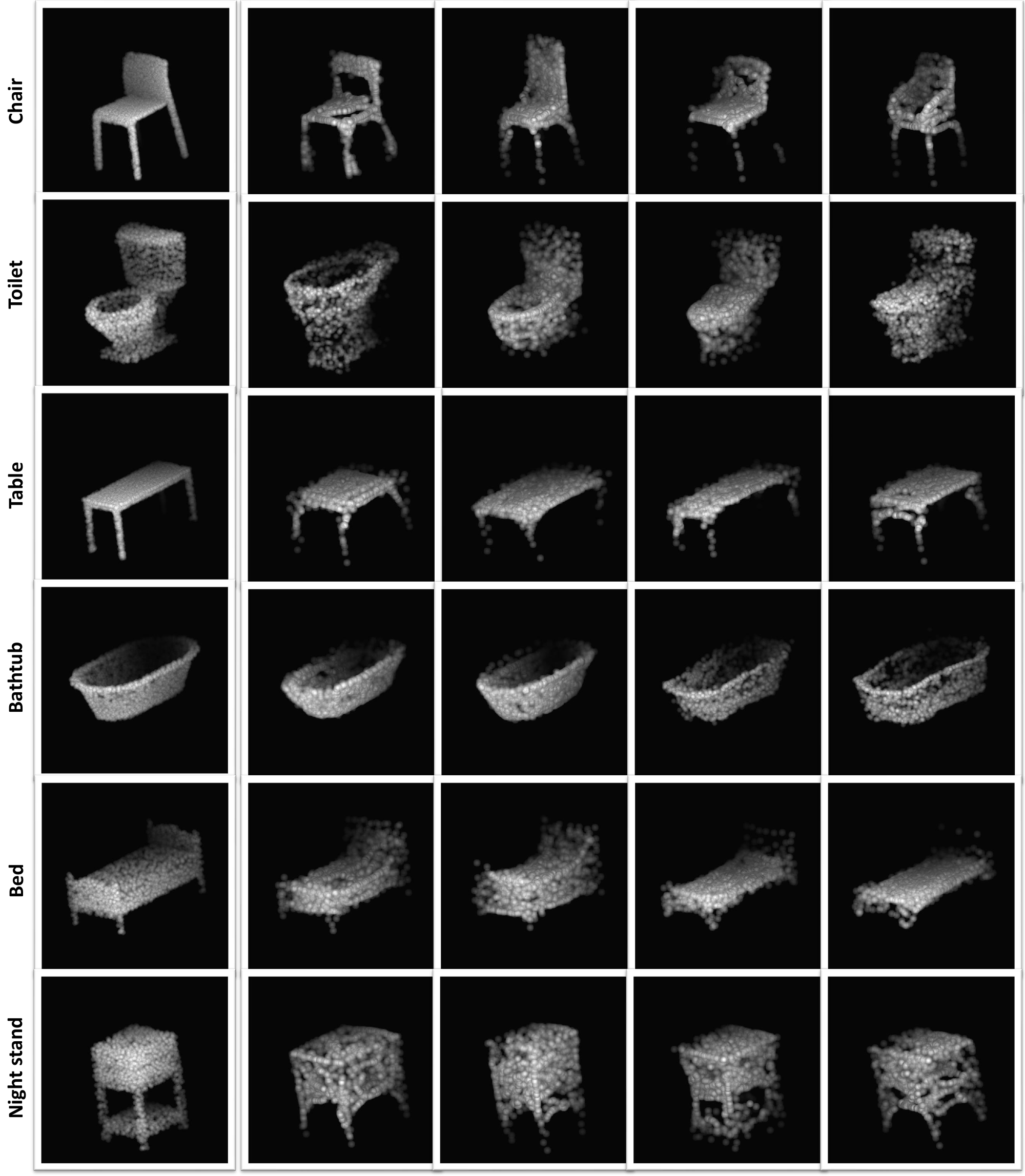}}\vspace{-5pt}
\caption{Sample point clouds generated by GDPNet. Each row corresponds to one category. The first column is a sample from ModelNet10 training set, and the rest of the columns are synthesized point clouds generated via SGLD.}
\label{fig:samples}
\end{figure}

\begin{figure}[t]
\centerline{
\includegraphics[width = 8cm]{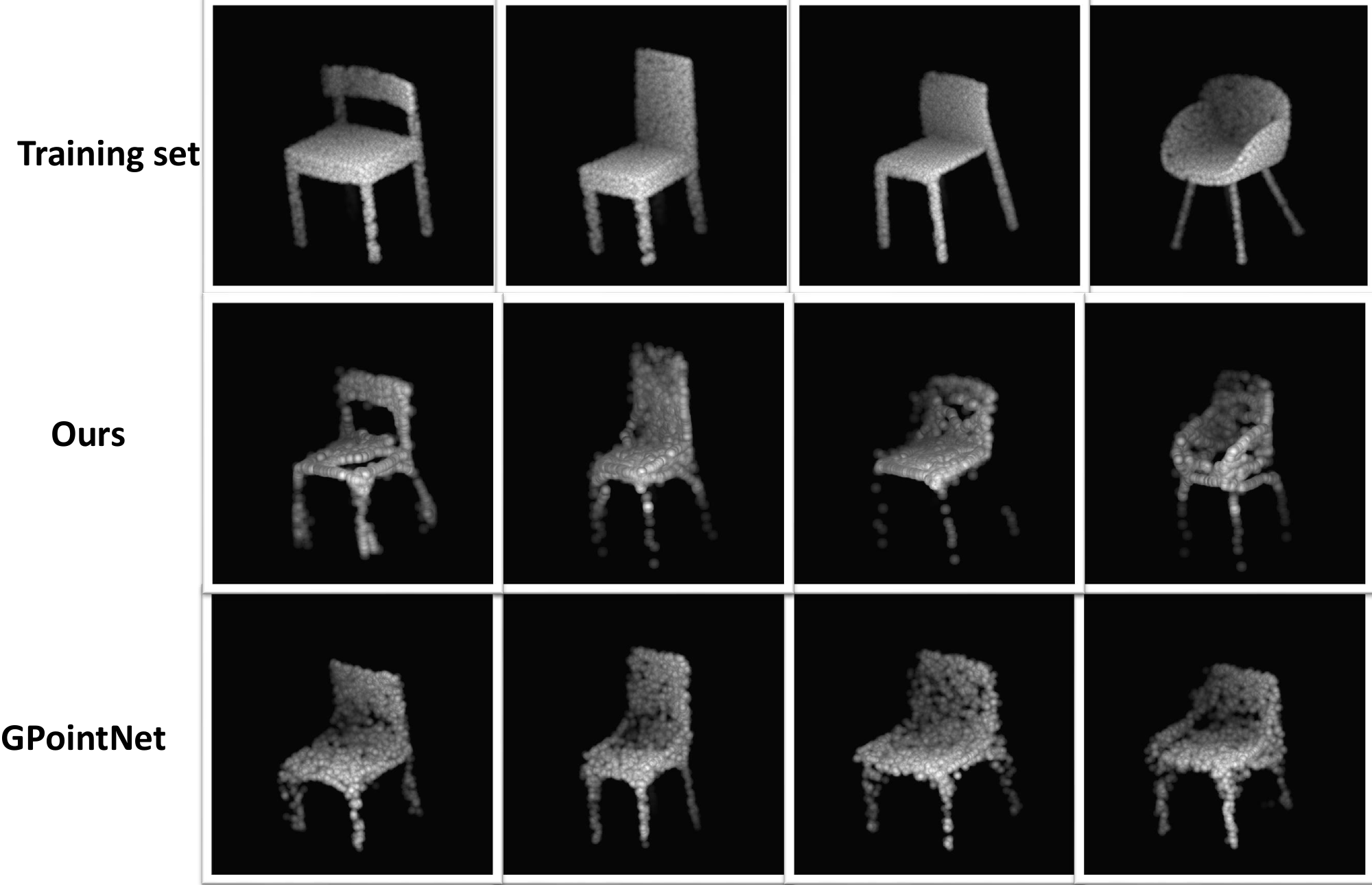}}\vspace{-5pt}
\caption{Sample point clouds generated by GPointNet~\cite{xie2021generative} and GDPNet. Our GDPNet generates chairs with more diverse styles, while GPointNet generates chairs with better details on the four legs.}
\label{fig:compair}

\end{figure}


\subsection{Results of Point Cloud Generation}

We compare the generative performance of GDPNet with four baseline generative approaches: l-GAN~\cite{achlioptas2018learning}, r-GAN~\cite{achlioptas2018learning}, PointFlow~\cite{yang2019pointflow} and GPointNet~\cite{xie2021generative}, and the results are reported in Table~\ref{tab:generation}. It can be observed that GDPNet achieves a competitive generative performance as compared to the state-of-the-art results of PointFlow and GPointNet even though our method only employs a single network to generate point clouds from 10 different categories of ModelNet10~\footnote{Let alone our GDPNet can also classify point cloud directly with an accuracy of $92.8\%$.}. For the generation of "monitor", GDPNet achieves an even better result than that of GPointNet, while being competitive with GPointNet for the rest of the categories. As shown in Figure~\ref{fig:samples_all}, GPDNet can learn the complex distributions among all the categories and generate point clouds of each category with diverse styles. It also can generate point clouds that have features from multiple categories, such as a toilet-like chair. This is because toilet and chair share some similar features, and GDPNet can generate samples that interpolate between them.


Figure~\ref{fig:samples} provides more sample point clouds generated by GDPNet for categories of "chair", "toilet", "table", "bathtub", "bed" and "night stand". These samples are selected when the GDPNet classifier has a classification confidence over 90\%. Not surprisingly, the generated samples for each category are exactly as GDPNet predicted, which also indicates an accurate classification of GDPNet. The results in Figure~\ref{fig:samples} show that GDPNet can learn the complex point distribution to generate quality point cloud samples. For "chair" and "table", the details of four legs are captured by our model, and the "bathtub" samples are as good as training samples, while the toilet samples are also decent. As for "night stand", whose shape is much more complex, the generated samples can still capture parts of the object features. We also used the official GPointNet checkpoint for the chair category to generate 1,000 samples, from which we selected some high-quality ones and compare them with the samples from our GDPNet. It can be observed from Figure~\ref{fig:compair} that GDPNet can generate chairs with more diverse styles, while GPointNet generates chairs with better details on the four legs. As a result, GDPNet has a slightly lower evaluation value on "chair" as reported in Table~\ref{tab:generation}.

\begin{table*}[t]
	\small
	\centering
	\begin{tabular}{|l|l|c|cc|cc|}
		\hline
		\multirow{2}{*}{\rotatebox{90}{}}  & \multirow{2}{*}{Method}   &  \multirow{2}{*}{JSD ($\downarrow$)}  & \multicolumn{2}{c|}{MMD ($\downarrow$)}        & \multicolumn{2}{c|}{Coverage ($\uparrow$)}    \\ \cline{4-7} 
		&                       &             & CD            & EMD            & CD             & EMD            \\\hline \hline

        \multirow{3}{*}{{Night Stand}} 
        
                                    & ReLU        & 0.487            & 0.911            & 1.320            & 43.02          & 47.67          \\
                                    & CELU    & \textbf{0.368} & \textbf{0.867} & \textbf{1.311} & 53.48 & 54.65             \\
                                    & CELU + SAM         & 0.377            & 0.886   & 1.369   & \textbf{54.83} & \textbf{59.30} \\ \hline

        \multirow{3}{*}{{Toilet}} 
        
                                    & ReLU        & 0.581            & 1.022            & 1.478   & 31.00          & 42.00  \\
                                    & CELU    & \textbf{0.546} & 0.941 & 1.380 & 32.00 & 37.00             \\
                                    & CELU + SAM         & 0.578            & \textbf{0.867}   & \textbf{1.314}   & \textbf{43.00} & \textbf{42.00} \\ \hline

         \multirow{3}{*}{{Monitor}} 
        
                                    & ReLU        & 0.410           & 0.692            & 1.261   & 45.00          & 46.00  \\
                                    & CELU    & \textbf{0.389} & 0.571 & 1.085 & 46.00 & 44.00             \\
                                    & CELU + SAM         & 0.434  & \textbf{0.535} & \textbf{1.029} & \textbf{52.00}             & \textbf{46.00}  \\ \hline

         \multirow{3}{*}{{Chair}} 
        
                                    & ReLU        & 0.449            & 0.916            & 1.495   & 44.00          & 46.00  \\
                                    & CeLU    & \textbf{0.363} & \textbf{0.8541} & \textbf{1.316} & \textbf{45.00} & \textbf{51.00} \\
                                    & CELU + SAM   & 0.387  & 0.909 & 1.361 & 44.00    & 50.00 \\ \hline
	\end{tabular}\vspace{-5pt}
	\caption{The impacts of SAM and activation functions on the qualities of generated point clouds by GDPNet. $\downarrow$: the lower the better; $\uparrow$: the higher the better. MMD-CD scores are multiplied by 100; MMD-EMD scores and JSDs are multiplied by 10.}
	\label{tab:comparison}  
\end{table*}

As for model complexity, GDPNet has a 1/10 model size of GPointNet since GDPNet only trains one single network to generate point clouds for all the 10 categories, which is a significant advantage of our method as compared to other generative approaches. For models based on VAEs or GANs, their model sizes are even larger than that of GPointNet as they need auxiliary networks for training. 

\subsection{Ablation Studies}
We study the impacts of SAM and the activation functions on the performance of GDPNet for point cloud classification and generation, respectively.


\textbf{Point cloud generation} 
Table~\ref{tab:comparison} reports the impacts of SAM and activation functions to the generative performance of GDPNet. By replacing the popular ReLU activation function~\cite{relu10} with CELU~\cite{barron2017continuously}, a continuously differentiable exponential linear unit, GDPNet achieves notable quality gains in generating point clouds of different categories. This observation is consistent with that of Zhao et al.~\cite{coarse2fine2021ebm} who demonstrate that the smoothness of the activation functions substantially improves the SGLD sampling process and thus the synthesis quality. Table~\ref{tab:comparison} also shows that incorporating SAM to GDPNet does not improve the synthesis quality consistently. However, from our experiments, we find that SAM facilitates the convergence of the model, stabilizes the training of GDPNet, and improves the classification accuracy as shown in the experiments below.  

\textbf{Point cloud classification} Table~\ref{tab:classification} reports the impacts of SAM and the activation functions on the classification performance of GDPNet. It can be observed that without SAM and CELU activation function, GDPNet has a non-competitive classification accuracy of 90.7\% as compared to the standard PointNet classifier, which achieves a 92.8\% accuracy. Incorporating SAM into GDPNet significantly improves the classification accuracy (92.8\%), which matches with that of the standard PointNet classifier. Replacing CELU by ReLU in GDPNet does not affect the classification accuracy much (92.9\% vs. 92.8\%), but GDPNet with CELU achieves the best synthesis quality as shown in Table~\ref{tab:comparison}. Therefore, GDPNet with SAM and CELU bridges both the classification accuracy gap and the synthesis quality gap as compared to the standard PointNet classifier and state-of-the-art generative approaches.

It is worth mentioning that even though l-GAN~\cite{achlioptas2018learning}, PointFlow~\cite{yang2019pointflow} and GPointNet~\cite{xie2021generative} achieve better classification accuracies as reported in Table~\ref{tab:classification}. They can not classify point clouds directly since they are generative models. Specifically, to classify point clouds with GPointNet~\cite{xie2021generative}, an one-versus-all SVM classifier needs to be trained on the extracted features of GPointNet and the class labels. A similar procedure has also been used by l-GAN and PointFlow for classification. In contrast, GDPNet does not need any extra training step for classification, which is another significant advantage of GDPNet as compared to these generative approaches.


\vspace{-5pt}
\begin{table}[!h]
\centering
\begin{tabular}{|l|c|}
			\hline
			Method    & Accuracy \\ \hline \hline

			l-GAN$^*$~\cite{achlioptas2018learning} & 95.4\% \\
			PointFlow$^*$~\cite{yang2019pointflow} & 93.7\% \\ 
			GPointNet$^*$~\cite{xie2021generative} & 93.7\% \\ \hline
   PointNet~\cite{qi2017pointnet}  & 92.8\% \\
   GDPNet w/ ReLU - SAM & 90.7\% \\
   GDPNet w/ ReLU & 92.9\%\\
   GDPNet - SAM & 90.3\% \\
   GDPNet & 92.8\% \\\hline
\end{tabular}\vspace{-5pt}
		\caption{Point cloud classification accuracies on ModelNet10. * denotes the method needs to train an SVM classifier on the extracted features for classification.}
		\label{tab:classification}
\end{table}

\vspace{-10pt}
\section{Conclusion}

This paper introduces GDPNet, a hybrid generative and discriminative model for point clouds, that is based on joint energy-based models. GDPNet further leverages SAM and CELU activation function to bridge the classification accuracy gap and the generation quality gap to the standard PointNet classifier and state-of-the-art generative models. Compared to prior generative models of point clouds, GDPNet only trains a single compact network to classify and generate point clouds of all categories. Experiments demonstrate our GDPNet retains strong discriminative power of modern PointNet classifiers, while generating point cloud samples rivaling state-of-the-art generative approaches. To the best of our knowledge, GDPNet is the first hybrid generative and discriminative model for point clouds. 

{
    \small
    \bibliographystyle{ieeenat_fullname}
    \bibliography{egbib}
}


\end{document}